\documentclass[11pt]{article}
\usepackage[table,xcdraw]{xcolor}

\usepackage[final]{ACL2023}
\usepackage[utf8]{inputenc}
\usepackage[T1]{fontenc}
\usepackage{booktabs}
\usepackage{enumerate}
\usepackage{graphicx}
\usepackage{listings} 
\usepackage{longtable}
\usepackage{multirow}
\usepackage{subcaption}
\usepackage{textcomp}
\usepackage{times}
\usepackage{url}
\usepackage{inconsolata}
\usepackage{microtype}
\usepackage{latexsym}
\usepackage{array,multirow}
\usepackage{comment}
\usepackage{dingbat}
\usepackage{algorithm2e}

\usepackage{amssymb}
\usepackage{amsmath}
\usepackage{graphics}
\usepackage{paralist}
\usepackage{comment}

\hypersetup{pdfauthor={Blake (Ralph) Vente},pdftitle={Inroads to A Structured Data - Natural Language Bijection and the role of LLM annotation}}

\usepackage{geometry}

\title{
Inroads to a Structured Data $\leftrightarrow$ Natural Language Bijection\\
and the role of LLM annotation 
}

\author{\stepcounter{footnote}Blake Vente\thanks{\;\;This work was completed as Graduate Coursework over the course of one semester with revisions after submission. Special thanks to the teaching staff of Columbia University's F2023 Natural Language Generation and Summarization course. All errors in this work are my own. } \\ rv2459@columbia.edu.} 

\begin{document}
\maketitle
\begin{abstract}
This work finds limited evidence supporting the theory that using multiple tasks with sequence-to-sequence transformer language models can improve performance on some metrics. In particular, the multi-task generalist t5-small outperforms the specialist t5-small with a $F_1$ of $0.771$ up from $0.692$, which may point to underlying cross-task knowledge generalization. This further suggests that even with the same network, "re-using" the same data in a different way may lead to higher performance in some metrics. However, the inverse task alone is likely only an optimization strategy, since it does not yield a significant general improvement at the model sizes explored in this work. Also, adding $\approx 4500$ LLM annotated records (interlaced with the $12800$ WebNLG training records) does not substantially change automatic metric performance compared to the same t5-small model without the synthetic data. This may be due to a learning capacity bottleneck on account of model size, and decreases observed may be due to distributional differences in the corpora. Future research using larger models or human evaluation is required to more fully explain the mechanisms contributing to performance on these tasks.
\end{abstract}

\section{Introduction}

\subsubsection{Motivation} It would be an understatement to say there has been an explosion in interest in Large Language Models (LLMs) for assisting with knowledge work. At the same time, we are grappling with the "hallucination" problem  \cite{gabriel-etal-2021-go, kryscinski-etal-2020-evaluating}. To address this, have proposed Retrieval-Augmented generation (RAG), placing unstructured documents into the context windows of Large Language Models. Still, there is a relative dearth of researching structured queries for RAG from structured sources compared to unstructured sources \cite{li2022survey, shuster-etal-2021-retrieval-augmentation}. It may one day be possible for smaller language models to match or best larger models for factual data recall tasks by iteratively querying databases of facts with their sources. This work is a first step towards small pre-trained Language Models (PLMs) which add structure to text documents. Further aspirational use-cases follow in Appendix \ref{usecase}.\footnote{The repository can be found at \url{https://github.com/rvente/nlgs-research/}, which contains links to published model weights.}

\section{Related Work}
    
\paragraph{Multitask Training}  Given sufficient learning capacity, it is possible for a language model to score better on all tasks by increasing the number of tasks via a mechanism called "co-training transfer." \cite{DBLP:journals/corr/abs-2111-10952}. The authors present ExT5 with the gargantuan figure of 107 training tasks and evaluate how and when a multi-tasking model can outperform a model of the same size with no additional data from that particular dataset. This work inspired the choice of WikiBio as an appropriate related task.

\paragraph{Layer Freezing Policy} The current state-of-the-art methodology ``control prefixes''  reaches a BLEU score of  .67 on seen entities and a .61 overall \citeauthor{clive2022control}.  ``Control prefixes'' are control signals that are directly appended to the hidden states of the network to guide generation while using a frozen pre-trained language model (PLM). This work recommends against freezing any layers for the propagation of prompt prefixes in control \citet{raffel2022exploring}, while the context is different, this work also does not employ layer freezing.

\paragraph{Semantic Parsing} For Semantic Parsing (sentence-to-data and s2d used as direct synonyms in this work), the current state of the art on the WebNLG+ dataset \citet{dognin2021regen} with $F_1$ score of .723. This work very innovatively frames text generation as a multi-step decision process and discusses adaptations to use non-differentiable evaluation metrics as a reinforcement penalty to guide generation. For this task, $F_1$ score is derived from framing semantic parsing as a classification task outputting an unordered set of RDF triples. This WebNLG+ corpus was originally released as part of a competition, and the work \citet{castro-ferreira-etal-2020-2020} compiles the results and summaries of many approaches. Notably, there is evidence for multitask learning as an approach with the bt5 network presented in the competition. In particular, \citet{agarwal-etal-2020-machine} used cross-lingual multitasking for English and Russian, with a final Resulting $F_1$ score of .877.  By contrast, this work's multi-tasking is simply learning both transformations, from structured data to natural language sentences, and the inverse transformation.

\paragraph{Synthetic Data Generation}
According to research by \citeauthor{axelsson2023using} instruction-tuned LLMs perform data-to-sentence generation but despite high fluency, ChatGPT earns a paltry 0.424 BLEU. To me, this speaks to the limitations of automatic evaluation. Work by \citet{shin2022fewshot} shows that using longer beam lengths increases accuracy on two d2s corpora, Overnight and SMCalFlow. Separately, work by \cite{tang2023does} shows promising results from generating synthetic annotations for the BioCreative VCDR corpus of clinical text ($F_1$ from 0.2337 to .6399 for named entity recognition; and $F_1$ from 0.7586 to 0.8359 percent for relation extraction). Finally \citet{hsieh2023distilling} uses PaLM-generated rationales as a proxy objective for smaller pretrained language models whose primary task is natural language inference on the Stanford Natural Language Inference (SNLI) and Adversarial Natural Language Inference (ANLI) corpora. Since there is some evidence that synthetic data can enhance performance, this work also incorporates synthetic data generation.

\section{Data}

\begin{figure*}[h]
    \begin{subfigure}{0.48\textwidth}
    \vspace{1em}
    \centering
\begin{tabular}{llll}
\toprule
     & \multicolumn{2}{c}{WebNLG} & WikiBio         \\
     & Sentence      & RDF        &  Sentence \\
     \midrule 
mean  & 310.68        & 139.54     & 526.48          \\
std   & 171.69        & 83.79      & 593.25          \\
min   & 22.00         & 22.00      & 6.00            \\
25\%  & 174.00        & 71.00      & 183.00          \\
50\%  & 291.00        & 130.00     & 318.00          \\
75\%  & 421.00        & 190.00     & 627.00          \\
max   & 1191.00       & 657.00     & 56698.00       \\
\bottomrule
\end{tabular}
\caption{Length in characters for the various corpora. These statistics are reported before any preprocessing is done to the text. For edit distance, lower scores indicate better performance}\label{corpusstats}
\end{subfigure}
    \begin{subfigure}{0.48\textwidth}
\centering
\begin{tabular}{lrr}
\toprule
Corpus &  WebNLG &  WikiBio \\
\midrule
train set &   12876 &   582659 \\
validation set &    1619 &    72831 \\
test set &    1600 &    72831 \\
$\Sigma$ &   16095 &   728321 \\
\bottomrule
\end{tabular}
\caption{Number of records for each corpus by partition set.}
\label{tab:splits}
\end{subfigure}
\begin{subfigure}{0.48\textwidth}
    \centering
    \centerline{
    \includegraphics[width=8cm]{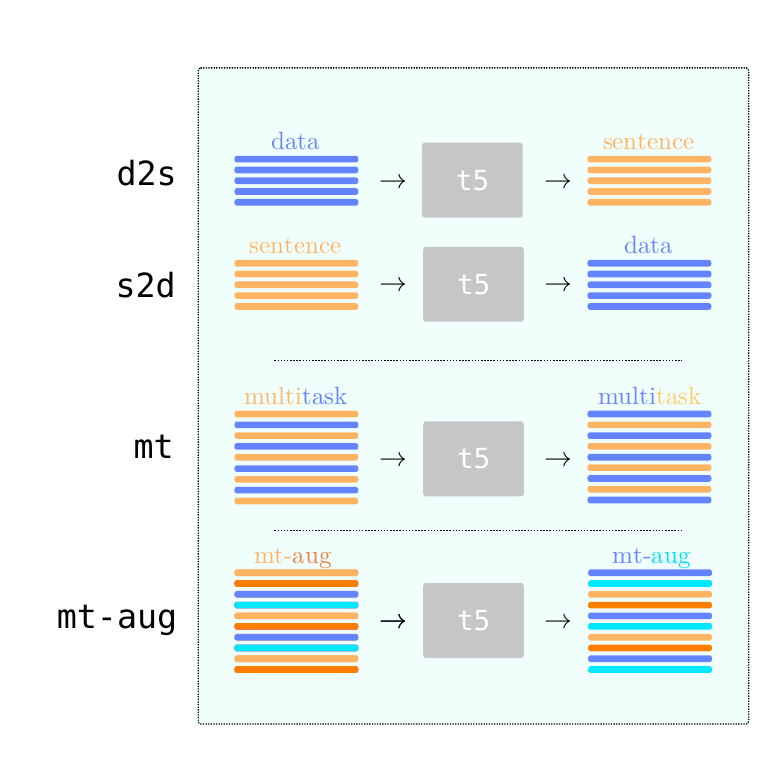}
    }
    \caption{\textbf{Model Training Scheme} There are three treatments are: (1) specialists: data-to-sentence and sentence-to-data (2) multi-task and (3) synthetic data-augmented variant of 2.}
    \label{traininscheme}
\end{subfigure}
\begin{subfigure}{0.48\textwidth}
    \centering
    \centerline{
    \includegraphics[width=7cm]{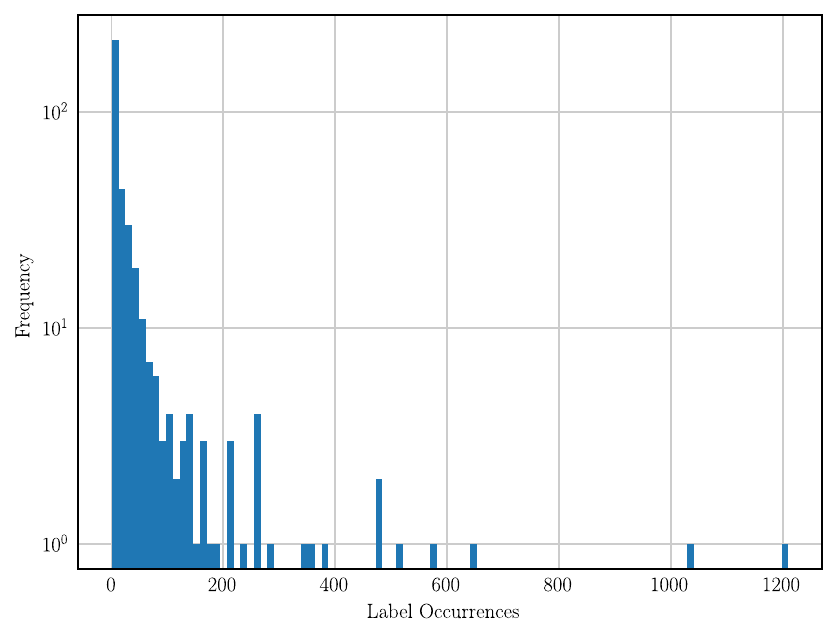}
    }
    \caption{Some relations such as \texttt{country} and \texttt{location} have more than a thousand occurrences. But there is a long tail of relations that only occur once, including \texttt{nearestCity} and \texttt{currentTeam}.}
    \label{fig:labelfrequency}
    
\end{subfigure}
    \caption{Corpus details and Training Scheme}
\end{figure*}

The primary corpus for this work is the ``bi-directional WebNLG+'' (also called WebNLG+ 2.0 or WebNLG2020) variant of the widely-used WebNLG corpus introduced in \citet{gardent-etal-2017-creating}. The ``bi-directional'' descriptor alludes to the two sub-tasks: RDF-to-sentence and sentence-to-RDF. \citet{castro-ferreira-etal-2021-2020} calls these tasks generation (data-to-sentence, d2s) and semantic parsing (sentence-to-data, s2d) respectively. RDF triples contain three terms, with entities on either side surrounding a relation. The relations themselves have a highly imbalanced distribution, with most relations occurring extremely rarely and few relations occurring extremely frequently. This can be seen in Figure~\ref{fig:labelfrequency}.

Examples of individual training examples can be found in Figure~\ref{fig:dataexamples}. The average length in tokens can be found in \ref{corpusstats}. Each record of the corpus has exactly one set of RDF triples and on average 2.66 of natural language translations of that particular triple set.  In turn, RDF triple set has an average of 2.9 individual RDF triple items with a standard deviation of 1.5. This corpus has 12,876 total records in the training set, 1619 in validation, and 1600 in the test set. The sentences of Wikibio are much longer on average: with 526.48 characters per sentence.

The d2s task inputs RDF triples and outputs natural language text. Conversely, the s2d task inputs natural language sentences and outputs structured data. This work uses these terms as direct synonyms of generation and semantic parsing respectively. Basic data cleaning was performed on the corpus, including Unicode to ASCII remapping.\footnote{Substituting spaces in place of underscores caused the length in tokens to fall from about 54 to  41, representing a 24 percent savings}. 

\section{Methods}

This work establishes baseline models, referred to as the ``specialists'' that are fine-tuned on a single task and compares it to the multi-taskers, ``generalists''.  To this end, this work fine-tunes pre-trained versions of the t5 series from \citet{su-etal-2021-plan-generate} three sub-tasks: data-to-sentence, sentence-to-data, and multi-tasking. The source model is found at HuggingFace\footnote{\url{https://huggingface.co/t5-small}}. For the multitasking model, an arbitrarily-designed control prompt specifying the task is prepended to the input, \verb|d2t 0:| and \verb|t2d 1:|. The single-task variants do not need this additional information to specify the task. The training tasks are interlaced, namely, the tasks alternate ABAB until all training examples are exhausted. For all tasks, structured data must be serialized for insertion into the context window. For data-to-sentence, the RDF triple terms are joined with vertical bars \verb+|+, and each triple ends with a \verb|;|. For sentence-to-data, the natural language sentence is taken as input and decoded into the serialized RDF representation.

This work also incorporates Google's PaLM Large Language Model \cite{chowdhery2022palm} for automatic data annotation. This work uses the text-bison version of Google PaLM 2 via Google Cloud. PaLM is an instruction-tuned large language model. Due to time constraints, I leave it to future work to rigorously evaluate PaLM on the sentence-to-data task in WebNLG. This work only uses the LLM as an annotator, outputting WebNLG-style triples for each WikiBio record sent to it.

\section{Experiments}

The Huggingface wrapper to the reproducible sacreBLEU library from \cite{post-2018-call} computes BLEU-4 scores as specified by in \cite{papineni-etal-2002-bleu}. Comparison was case-insensitive. Huggingface's \texttt{evaluate} library computes BERTScore \cite{zhang2020bertscore} with \texttt{distilbert-base-uncased}, a distilled variant of the English Bert model. Likewise, \texttt{evaluate} computes the RougeL score, the Longest Common Subsequence Rouge Score (denoted RougeL), wrapping the Google Open Source implementation of Rouge.

The experiments adhered to the standard train, development, and test splits of the corpus. As the baseline model, this work fine-tunes the Huggingface \texttt{t5-small} and \texttt{t5-base} networks with an effective batch size of 64 and 32 respectively. Both networks were trained with a learning rate of $2 \cdot 10 ^ {-4}$ for 5 epochs. The shortest training time for any network was about 30 minutes and the longest took about 1.5 hours (not counting evaluation). There was a wide variance of decoding time during the predictions in the validation set (more on this in Figure~\ref{palatul}). Decoding was performed with 4 beams, temperature was reduced from the default of 1.0 to 0.9, and top-$k$ filtering was kept at the default of 50 tokens. The first treatment is the multitask training objective, where the model must predict perform s2d and d2s alternating in the records, and the second treatment is the same as the first, but also incorporates LLM annotated WikiBio data randomly inserted into the training set, as denoted in \ref{traininscheme}.

Training proceeded on a 24 GB Nvidia RTX 3090. At the time of training, t5-base with a batch size of 32, \texttt{nvidia-smi} reports 19289 mebibytes (MiB; \ensuremath{\approx} 20.2 gigabytes) of Video RAM (VRAM) in use. This figure slowly grows as memory fragmentation occurs during the training process. For example, by epoch 1.4 it grew by 1500 MiB which is about 8 percent of VRAM consumption at the start of training. The amount of memory leaking is linear. Decreasing batch size to 16 resulted in a divergence of training loss within 5 epochs on t5-base. To resolve this, I used \verb|gradient_accumulation_steps| for a larger effective batch size even with a smaller in-memory batch size \footnote{\url{https://huggingface.co/docs/accelerate/usage_guides/gradient_accumulation}}.

\section{Results}

\begin{figure*}
\centering

\begin{tabular}{@{}r|ccccc|@{}}\cmidrule(l){2-6}
                                 & BLEU$\uparrow$  & BERTScore $\uparrow$  & RougeL$\uparrow$  & $F_1$$\uparrow$             & Edit Distance $\downarrow$          \\ \cmidrule(l){2-6} 
\multirow{-2}{*}{}               & \multicolumn{3}{c|}{data-to-sentence} & \multicolumn{2}{c|}{sentence-to-data} \\ \cmidrule(l){2-6} 
\rowcolor[HTML]{ECF4FF} 
t5-small                         & 0.641 & 0.953        & 0.745     & \multicolumn{1}{|c}{0.692}         & 16.854                \\
t5-base                          & 0.671 & 0.957        & 0.767     & \multicolumn{1}{|c}{0.928}         & 15.489                \\ \cmidrule(l){2-6} 
\multicolumn{1}{r|}{}            & \multicolumn{5}{c|}{multi-task}                                               \\ \cmidrule(l){2-6} 
\rowcolor[HTML]{ECF4FF} 
\cellcolor[HTML]{ECF4FF}t5-small & 0.618 & 0.949       & 0.733     & 0.771         & 16.16            \\
t5-base                          & 0.602    &  0.945        & 0.718     & 0.887         & 15.711	               \\ \cmidrule(l){2-6}
\multicolumn{1}{r|}{}            & \multicolumn{5}{c|}{multi-task llm-augmented}                                     \\ \cmidrule(l){2-6} 
\rowcolor[HTML]{ECF4FF} 
t5-small & 0.610 & 0.948 & 0.730 & 0.754 & 16.359            \\ \cmidrule(l){2-6}
\end{tabular}
\caption{This table reports the results of the six experiments run on the WebNLG corpus. The identical standard test set was used for all evaluation $N=1600$. The figures reported are not directly comparable to prior work due to cleaning and preprocessing.}
\end{figure*}

In this case, taking t5-small  and training it with the multi-task objective caused it to perform higher on the d2s metrics $F_1$ (from 0.692 to 0.771) and Edit Distance (from 16.854 characters to 16.16), compared to the same model trained on just a single objective. At the same time, it didn't substantially change BERTScore, and resulted in a slight decrease in Rouge (from 0.745 to 0.733), and BLEU (from 0.641 to 0.618). This might show signs of \textit{multi-task generalization}, but only in one direction. One conjecture into the underlying mechanism could be that the model learns vocabulary that is re-used when computing labels. This would explain why the converse isn't true: RDF label vocabulary is a subset of the text vocabulary.

However, in these same circumstances, the multi-task t5-base model was lower than the respective specialist models in every metric. Its closest was in BERTScore where performance only fell slightly (down to 0.945  from 0.957). Taken on its face, this might suggest there exists an ideal ratio between training set size, fine-tuning size, and model size similar in spirit to Palm's compute optimal scaling hypothesis \cite{hoffmann2022training}. However, much more research is needed to add confidence to this finding, explored more in the Limitations portion.

As an aside, pre-pending the task may not be strictly necessary because the model was able to identify the task automatically, probably from the vertical bars that delimit the RDF triples. 

\paragraph{LLM Annotation} 5,000 annotations were requested from PaLM using the text-bison API. About 400 were content-filtered: by default, the API filters out model outputs about protected groups. About 100 contained malformed expressions.\footnote{To be space-efficient, my prompt is available in my code repository under the \texttt{palm} folder. The pickle file for the annotated records is \texttt{wikibio\_llm\_annot.pkl}} The 4543 remaining records were processed in the same manner as the WebNLG corpus. Qualitatively, the model appeared adequate to perform semantic parsing, and in the samples I have observed, did not add any information not originally present in the prompt. However, t5-small-aug did not perform better than t5-small. In fact, the augmented variant performed slightly worse in all tasks. Perhaps this was due to not enough learning capacity in the t5-small base model. Perhaps idioms of WikiBio were different enough as to slightly harm performance. More work is needed for a full explanation. 

\section{Error Analysis}

``Repetition loops'' as in \cite{xu2022learning} refers to when the model repeats the same few sentences over and over again. I observed this ``hallucination'' occur, even in the largest set of models trained \texttt{t5-base-mt}. Curiously, it was isolated to a particular word Palatul, Romainan for "palace" \ref{palatul}. This behavior diminished but was present in some of the trained models nonetheless. One trained model had a ``Palatul'' glitch token. In Figure \ref{palatul}, each row is an example of t5-base-mt falling into repetitive generation cycles even when the \verb|no_repeat_ngrams| parameter is set to 3. Ultimately it is the parameter of \verb|max_length| that terminated generation. To reduce validation length, one can set this value low.

The errors show the existence of some ``false penalties'' where the model's outputs are sensible but not recognized by automatic evaluation. As seen in Figure \ref{badbleu}, BLEU does not take into account sentence variations or semantic similarity. In those records (from d2s-t5-small) the BLEU was than 0.15 In record 222, we can see a penalty for using 30.0 g instead of 30 grams. We also see penalties from differential placement of relative clauses (in this case "whose", "who is from spain"). Across these samples, the BERTScore is 0.893. This casts further doubt into these forms of automatic evaluation.

The plot in Figure \ref{s2dperf} document on the worst performers on semantic parsing by F-measure. In all variants after normalizing by the prevalence in the training set, I can confirm that \verb|SportsTeam| was disproportionately difficult for the model, even though these samples were shorter than average. Due to time constraints, it can be left to future work to investigate further, especially the question of t5-base-aug. 

\section{Conclusions, Limitations, and Future Work}

The key finding of this work is that even with no additional data or an increasing model size, learning the inverse task may increase performance for some sequence-to-sequence sub-tasks. This finding is significant because it upholds the value of multi-tasking, even in a two-task context with no additional data. But this is not a generally reliable method as only some metrics increased while others decreased. Scaling up the model was a more reliable way to gain more performance, but once scaling up shows diminishing returns, this method may be used as a last step optimization.

\paragraph{Limitations} In this work, every model was trained with the same control value of 5 epochs through the training set, so every model trained on the same number of records the same number of times. However, future work should certainly take into consideration the effect of training each and every network to the same degree of convergence instead. Experimenting with different training hyper-parameters may also prove fruitful. In this work, no statistical significance tests were performed - given more time, I would train each network multiple times, or use several independently initialized generations for test set evaluation.

In this work, long entity names were negatively correlated to performance on both sub-tasks. Since the task is framed as a sequence-to-sequence problem, the individual decoding errors add up. It's possible to "compress" the text form of the entity so that long entity names don't "distract" from the true purpose of the model. In particular, very long entity names could be bound to separate short-names and then unbound later. To illustrate one intuitive compression scheme Appendix \ref{compression} shows an example. One can see that this saves on token count compared to repeating the entity name without losing any information. However, it may require the network to learn to bind a name to a variable and use it appropriately. In this work, this optimization was not explored, but future work may consider this.

The "Palatul" issue shows that the "Hallucination problem" still requires addressing. I conjecture that one approach to mitigate this is defining a task to the index of the spans themselves for entity names, (just as an image segmentation model might predict bounding boxes around objects in an image). By forcing a model to work with indices directly, extracting a sequence not present in the input would be structurally impossible. Prior work shows that this method is feasible in principle \cite{subramanian-etal-2021-spanpredict}, but much further research is required. To be specific, instead of the probabilistic mechanism of decoding input tokens from encoded representations through the network, the task would entail directly predicting the index of the substrings that comprise the terms of each RDF triple.

There is still no defacto normal meaning-representation (sememe) for all words, so "Footballer" and "Soccer player" may have different RDF labels, but this isn't desirable in general for automatic evaluation as seen in \ref{qualeval}. Even an informal notion of \textit{bijection} between Natural Language and Structured data demands solutions to the ontology problem. It could be a daunting task to develop a formal ontology for all RDF triples, but doing so would bring more confidence to the relevance of the $F_1$ scores used in the semantic parsing task. Future work may also consider human expert evaluators to evaluate the quality of the semantic parses of the output.

\bibliography{anthology,custom}
\bibliographystyle{acl_natbib}
\newpage{}

\onecolumn
\part*{Appendix}
\appendix{}

\section{$F_1$ measure definition}
For rigor's sake, $F_1$ score is not used in its usual sense, so it is fruitful to formally extend the definition of $F_1$ for sets of open-vocabulary labels. To my knowledge, there is no defacto implementation or pseudocode between all works reporting $F_1$ score for semantic parsing (s2d) task.
Let $P$ be the set containing predictions in the form of RDF sequences from the model.
Let $G$ be the set containing the ground-truth references in the same format.
Let $\textbf{hm}(a,b)$ be the harmonic mean of integers $a$ and $b$, and let $\textbf{hm}(0,x)=\textbf{hm}(x,0) =0$ for all $x$. And let $|X|$ denote the cardinality of set $X$.
\begin{algorithm}\label{algo}
\textit{\tiny integer} $\textsc{tp} \gets | P \cap G | $\;
\textit{\tiny integer} $\textsc{fp} \gets | P - G | $\;
\textit{\tiny integer} $\textsc{fn} \gets |G - P |$\;
\textit{\tiny decimal} $\textsc{prec} \gets \textsc{tp} / (\textsc{tp} + \textsc{fp} + \varepsilon)  $\;
\textit{\tiny decimal} $\textsc{recl} \gets \textsc{tp} / (\textsc{tp} + \textsc{fn} + \varepsilon)  $\;
\textit{\tiny decimal} $F_1 \gets \textbf{hm}(\textsc{prec}, \textsc{recl})  $\;
\end{algorithm}

\noindent
In the following code, $P$ is the first argument and $G$ is the second. The strings in the sets are always computed by case-insensitive, whitespace-insensitive match.
\begin{lstlisting}[basicstyle=\small, breaklines=true]
f_measure(set("a"), set('a')) == 1
f_measure(set("ab"), set('a')) == 2/3
f_measure(set() , set('a')) == 0
\end{lstlisting}

\section{Semantics Preserving Compression Scheme}\label{compression}

For an exaggerated illustrative example, in the WebNLG schema, it would currently be denoted as

\begin{lstlisting}[basicstyle=\small, breaklines=true]
Spirit_of_future_yet_to_come | appears in | A_Christmas_Carol
Spirit_of_future_yet_to_come | is a | fictional_character
Spirit_of_future_yet_to_come | is a | ghost
Spirit_of_future_yet_to_come | createdBy | Charles_Dickens
Spirit_of_future_yet_to_come | appearsBefore | Ebenezer_Scrooge
\end{lstlisting}

\noindent
Tokens can be saved binding and referring to names in the following or equivalent compresion scheme.

\begin{lstlisting}[basicstyle=\small, breaklines=true]
let A = "The Spirit of Christmas Yet To Come";
$A | appears in | A Christmas Carol;
$A | is a | fictional character;
$A | is a | ghost;
$A | created by | Charles Dickens;
$A | appears before | Ebenezer Scrooge;
\end{lstlisting}

\newpage

\section{Data Samples}
\subsection{Arbitrary WebNLG Records}
\begin{figure*}[h!]
\centering
\begin{lstlisting}[basicstyle=\small, breaklines=true]
241 In Mexico, the spoken language is Spanish.
    Spanish is the language spoken in Mexico.
    The language of Mexico is Spanish.
242 One of the languages used in the Philippines is Arabic.
    Arabic is a language spoken in the Philippines.
    One of the languages in Philippines is Arabic.
    Arabic is one of the languages spoken in the Philippines.
243 Shumai is a variation of the dish Siomay.
    Siomay and Shumai are variations of the same dish.
244 Native Americans in the United States are one of the ethnic groups of the country.

241 Mexico | language | Spanish_language
242 Philippines | language | Arabic
243 Siomay | dishVariation | Shumai
244 United_States | ethnicGroup | Native_Americans_in_the_United_States
\end{lstlisting}
\caption{Some arbitrarily chosen samples of the sentence form of the data (top) with the associated raw data in triple form (bottom). Each entry that appears on a new line is one of the valid options for the data-to-text task. This means that BLEU score will acknowledge each of the variants. The natural language text is above and each numbered line is paired with the structured RDF triples below.}
\end{figure*}

\subsection{Arbitrary WikiBio Text Records}
\begin{figure*}[h!]
\begin{lstlisting}[basicstyle=\small, breaklines=true]
john chubb -lrb- 1816 -- 1872 -rrb- , was an english locksmith and inventor . he wrote an important paper on locks and keys , and was awarded the telford medal .
mary kendall browne -lrb- june 3 , 1891 -- august 19 , 1971 -rrb- was the first american female professional tennis player , a world no. 1 amateur tennis player , and an amateur golfer . she was born in ventura county , california , united states .
nicholas phillip ebanks -lrb- born 27 june 1990 -rrb- is a caymanian footballer who plays as a defender . he has represented the cayman islands during the 2010 caribbean championship and world cup qualifying matches in 2011 .
warren archard luhning -lrb- born july 3 , 1975 -rrb- is a retired canadian professional ice hockey winger .
\end{lstlisting}
\caption{Further arbitrarily chosen samples from the text column of WikiBio. These are pre-tokenized and have ASCII encoding for individual characters. The whole sequence is lowercased and \texttt{-lrb-} denotes ``left round bracket''.}

    \label{fig:dataexamples}
\end{figure*}

\newpage
\section{Model Outputs}
\subsection{Qualitative Observations on False penalties}\label{qualeval}
\begin{compactenum}
    \item Transitive relations whose arguments are swapped are not captured
    \item Synonymy/ reasonable alternatives not accounted for
    \item Equivalent formulations not accounted for
    \item Units and formulation 
    \item Inconsistent schema/synonymy
\end{compactenum}

\noindent

P = predicted, A = actual; examples extracted from t5-small trained to 5 epochs

\begin{lstlisting}[basicstyle=\small, breaklines=true]
1 P: Christian Burns|associated band/associated musical artist|Andrew Rayel
  A: Andrew Rayel|associated band/associated musical artist|Christian Burns
  
2 P: California|stone|Benitoite
  A: California|gemstone|Benitoite
  
3 P: Al Kharaitiyat SC|league|Qatar Stars
  A: Al Kharaitiyat SC|position|Qatar Stars League
  
4 P: Andrews County Airport|runway length|896
  A: Andrews County Airport|runway length|896.0
  
5 P: Atlanta|leader name|Kasim Reed
  A: Atlanta|leader|Kasim Reed
  P: United States|leader name|Barack Obama
  A: United States|leader|Barack Obama
\end{lstlisting}

\subsection{Low BLEU scores}
\begin{lstlisting}[basicstyle=\tiny, breaklines=true]
198 | ["April O'Neil was created by Kevin Eastman.", "Kevin Eastman is the creator of April O'Neil."]
    > Kevin Eastman created April ONeil. 
222 | ['Barny cakes can be served in 30 gram sizes.', 'Serving size for the Barny cakes is 30.0g.', 'The serving size of Barny cakes is 30.0g.']
    > Barny cakes have a size of 30.0 g.
229 | ['Bionico can be varied by using cottage cheese.']
    > Cottage cheese is a variation of Bionico.
288 | ['Abdulsalami Abubakar ended his career on 1999-05-29.']
    > Abdulsalami Abubakar's career ended on 29th May 1999. 
310 | ['Allan Shivers started his career from January 21, 1947.']
    > Allan Shivers began his career on 21 January 1947. |
419 | ['Alan Frew is a rock musician, which includes fusion and Bhangra styles.', "Alan Frew's genre is Rock music of which bhangra is a fusion of rock.", "Alan Frews' musical genre is rock music and a type of rock music fusion is Bhangra."]
    > Alan Frew performs rock music which has a fusion genre called Bhangra.
438 | ['Al Anderson of NRBQ is a country musician in which genre the banjo features.', 'Al Anderson of NRBQ performs country music which is a genre of music which uses the banjo.', 'Al Anderson (NRBQ band) performs country music, in which the banjo is one of the instruments.']
    > Al Anderson, a member of the NRBQ band, performs country music. Banjo is a musical instrument of country music.
551 | ['Baked Alaska comes from the country of France and one of the ingredients is sponge cake.', 'Baked Alaska (France) uses sponge cake as an ingredient.', "France's Baked Alaska includes the ingredient, sponge cake."]
    > Sponge cake is an ingredient in Baked Alaska which is from France. 
570 | ['Bionico requires granola as one of its ingredients and can be found in Guadalajara.', 'Granola is a required ingredient of the Guadalajara regional dish, Bionico.', 'Bionico, which contains granola, can be found in Guadalajara.'] 
    > Granola is an ingredient in Bionico which comes from the Guadalajara region.
755 | ['Alan Martin, whose club is Motherwell FC, played for Accrington Stanley FC who have their ground in Accrington.', "Alan Martin's football club is Motherwell FC and he has also played for the Accrington based club Accrington Stanley."]
    > Alan Martin is a footballer for the Accrington Stanley F.C. club which is located in Accrington.
809 | ['The epoch date of 1097 Vicia, which had 1928 PC as its former date, is 2006.12.31. Vicia has a periapsis measurement of 279142000000.']
    > 1097 Vicia, formerly known as 1928 PC, has an epoch date of December 31st 2006. It has a periapsis of 279142000000.0. |
885 | ['Found in Mexico, the food, Bionico (with granola as an ingredient), is served at the dessert course.', 'Bionico is served as a dessert course. It is found in Mexico and requires granola as an ingredient.']
    > Bionico is a dessert from Mexico and contains granola.
\end{lstlisting}

\noindent A sample of BLEU score < 0.15 generations from d2s-t5-small. The Training set record id starts each listing, followed by a vertical bar, then the list of ground truth translations, and finally \texttt{>} begins the generation of the model. \label{badbleu}

\begin{figure*}
    \centering
    \hspace*{-1.5cm}\includegraphics[width=18cm]{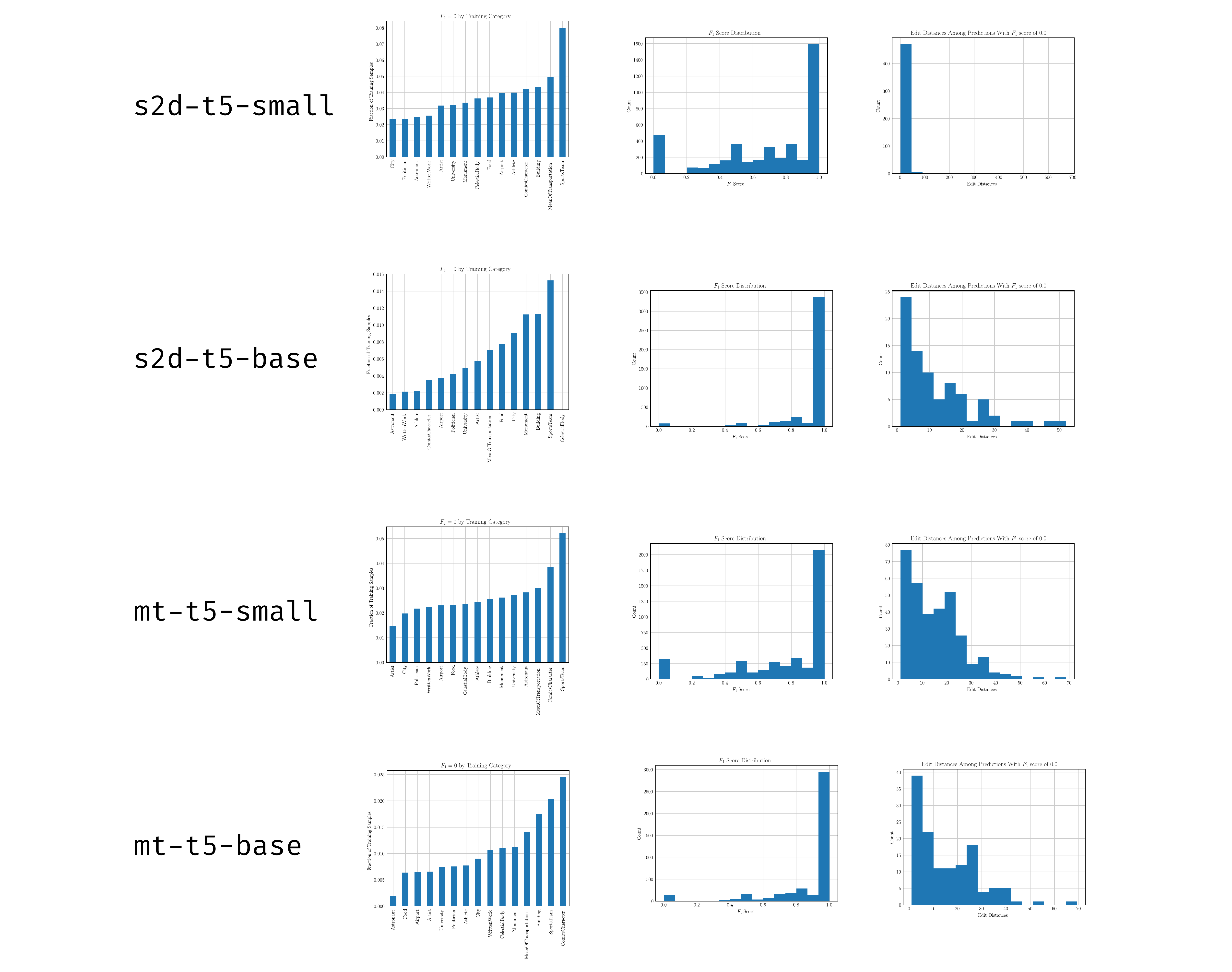}
    \caption{Performance on the sentence-to-data task by network size and training scheme. The first column represents the records that were awarded an $F_1$ of 0.0 in their training category. The second shows the overall $F_1$ score distributions. The final column shows the edit distance between the terms and their expected values. For the s2d-t5-small variant, the Edit Distance plot does not show it well, but the horizontal axis still included a single a single record that resulted from repetitive generation cycles. [vector images, the reader may zoom in for more detail]}
    \label{s2dperf}
\end{figure*}

\newpage

\section{Further Motivating Use-cases for Text-to-Data}\label{usecase}
To plan further developments to this task, future work may consider the following use-cases. All of these use-cases further substantiate the need for standard, normalized structured language formats (such as a formal ontology), which I consider the chief limitation faced by this task moving forward. 
\begin{enumerate}
    \item Pipeline generation for summarization: it's possible to break up a text into a set of facts and then rank the facts in terms of importance, de-duplicate them, sort them, and convert them back into a summary. Viewed this way, this holds promise for long-form summarization tasks.
    \item Summarization factuality evaluation: if there exists in the future, some exhaustive normal form capturing entities, relations between them, and their evolution over time, one way to evaluate factual fidelity would be to compute this ``factual decomposition.'' of the source text, compute the same for its summary, and then compute the degree of overlap. A faithful summary would be a proper subset of the facts in a source text and leave out only the least important details.
    \item Editing Wikipedia may be more beginner-friendly than editing on Wikidata. A system that extracts facts from new Wikipedia contributions and auto-syncs them with Wikidata (and the inverse) would be fruitful for the mission of keeping the two platforms in knowledge parity.
    \item Corroborating evidence and finding factual inconsistencies between various sources, which may be extended into automated fact-checking.
\end{enumerate}

\section{``Hallucination'' Example: The ``Palatul'' Glitch Token}
\begin{figure*}[h!]
\centering
\begin{tabular}{p{10cm} |l} \toprule predictions & palatul\_count \\ \midrule Austin is part of Hays County,Texas. & 2028 \\ Arlington is part of Tarrant County,Texas. & 2028 \\ The Acharya Institute of Technology & 2034 \\ The address of Amdavad ni Gufa is " & 2025 \\ alatul Ryan is the leader of the United States, where English is spoken. & 2025 \\ The ground of AFC Blackpool is located in Blackpool which is led by Gordon Marsden. & 2020 \\ Austin is part of Hays County,Texas,which has San Marcos as its county seat. & 2018 \\ Augustus Pugin was born in Bloomsbury, Ireland. He was the architect of Adare Manor, of Westminster, & 2013 \\ The 1 Decembrie 1918 University is located in Bucuresti Bucuresti Bucuresti Bucuresti Bucuresti Bucuresti Bucuresti Bucuresti Bucuresti Bucuresti Bucuresti Bucuresti Bucuresti Bucuresti Bucuresti Bucuresti Bucuresti Bucuresti Bucuresti Bucu & 2008 \\ \bottomrule \end{tabular}
\caption{A sample of the ``Palatul'' curse, generated from the \texttt{t5-base-mt} variant during training. Each string starts with a typical generation and diverges into this repetitive generation cycle. This behavior occurred with generation parameter \texttt{no\_repeat\_ngram=3}. I observed that values larger than this caused the model to backtrack for hours at certain points during training. This points  Future work might explore this problem in detail. This happened 9 times in the entire 1600 record test set during training of one model, and subsequently, the problem diminished without a clear cause. Similar behavior has been documented as "Glitch Tokens" \url{https://www.lesswrong.com/tag/glitch-tokens}}\label{palatul}
\end{figure*}

\end{document}